\documentclass[letterpaper, 10 pt, conference]{ieeeconf}  

\IEEEoverridecommandlockouts
\usepackage{graphicx} 
\usepackage{epsfig} 
\usepackage{amsmath} 
\usepackage{amssymb}  
\usepackage{algorithm,algorithmic}
\usepackage[margin=18mm]{geometry}
\usepackage{booktabs}
\usepackage{tabularx}
\usepackage{multirow}
\usepackage{makecell}
\usepackage{array}
\usepackage{adjustbox}
\usepackage{balance}
\usepackage{hyperref}

\newcolumntype{Y}{>{\centering\arraybackslash}X}
\title{\LARGE \bf
From Reach to Insert: Tactile-Augmented Precision Assembly under Sub-Millimeter Tolerances
}

\author{Xinpan Meng, Siyao Huang, JingPu Yang, Muyuan Ma, Zhenghua Ma, Lijun Han, \\Yuan Gao, Houcheng Li, and Long Cheng, \textit{Fellow, IEEE}%
\thanks{This work was supported in part by the Brain Science and Brain-like Intelligence Technology - National Science and Technology Major Project under Grant 2025ZD0215600, in part by the National Natural Science Foundation of China under Grant U25A20475, in part by the Beijing Municipal Natural Science Foundation under Grants F2024201068, L243014 and L232140, in part by Young Scientists Fund of The State Key Laboratory of Multimodal Artificial Intelligence Systems under Grant ES2P100114, and in part by the Fundamental Research Funds for the Central Universities. {\em (Corresponding author: Houcheng Li and Long Cheng.)}}
\thanks{Xinpan Meng, Houcheng Li, Muyuan Ma, Zhenghua Ma, Lijun Han, and Long Cheng are with the School of Artificial Intelligence, University of Chinese Academy of Sciences, Beijing 100049, China, and also with the State Key Laboratory of Multimodal Artificial Intelligence Systems, Institute of Automation, Chinese Academy of Sciences, Beijing 100190, China (email: long.cheng@ia.ac.cn).}%
}

\begin{document}
\maketitle
\thispagestyle{empty}
\pagestyle{empty}

\begin{abstract}

High-precision assembly frequently involves tight-tolerance insertions, where even slight pose errors can cause jamming or excessive interaction forces, making robust and safe insertion policies difficult to obtain. This paper proposes a tactile-augmented two-stage method that combines Imitation Learning (IL) and Reinforcement Learning (RL) for precision insertion tasks. In the first stage, IL learns a reaching policy with position generalization that grasps the peg and brings it to the vicinity of the target region. In the second stage, RL executes the insertion and enables recovery from failures during contact-rich interactions. To better exploit tactile feedback, we introduce tactile group sampling to increase coverage of critical contact segments during training, and design a tactile critic to more accurately evaluate policy values, improving insertion performance while maintaining low contact forces. We conduct systematic experiments across five hole geometries and three clearance settings. Results show that our method substantially improves insertion performance across all settings; under the most challenging 0.05\,mm clearance, it achieves a 67\% success rate while keeping contact forces low, reducing the maximum interaction force by 60\% and torque by 44\%, thereby validating both effectiveness and safety for precision assembly.

\end{abstract}

\section{INTRODUCTION}

In industrial assembly, precision insertion under tight tolerances is a key component of automated manufacturing, yet it remains highly challenging in real-world settings. Due to perception errors, control errors, and contact uncertainties, small discrepancies often arise between planned and executed trajectories. Under millimeter- to sub-millimeter-clearance peg-in-hole conditions, such small errors can be significantly amplified, leading to rapidly increasing interaction forces, jamming, and insertion failure. As a result, insertion success is strongly coupled with contact-force regulation, calling for safe, low-force strategies. Moreover, relying solely on vision or position feedback is often insufficient to infer contact modes and subtle misalignments, motivating the use of tactile feedback that directly reflects contact and force variations.
\begin{figure}[t]
	\centering
	\includegraphics[width=\linewidth]{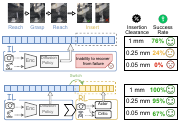} 
	\caption{Overview of the proposed two-stage tactile-guided assembly framework. 
		(a) \textbf{Reach Stage}: Vision-based imitation learning for autonomous grasping and approach. 
		(b) \textbf{Insert Stage}: Tactile-rich reinforcement learning for fine-grained alignment and insertion under tight tolerances ($0.05\,\mathrm{mm}$). 
		The method achieves high success rates across diverse geometries while effectively suppressing interaction forces.}
	\label{fig:teaser}
\end{figure}
Unlike many prior settings that learn insertion starting from a peg that is already stably grasped and initially aligned, a more realistic assembly pipeline is considered in which the robot must first grasp a randomly placed peg and then perform insertion under tight tolerances. The initial pose uncertainty of the peg and the in-hand pose error introduced during grasping accumulate over time and are further amplified during contact, making the insertion process more prone to eccentric contacts, increased torques, and jamming. This setting therefore imposes higher requirements on policy position generalization, alignment accuracy, and failure-recovery capability. Existing approaches can be broadly categorized into model-based methods, imitation learning (IL), and reinforcement learning (RL). Analytic methods, such as force control, impedance control, and search-based strategies, rely on accurate modeling of specific assembly scenarios and often lose robustness when hole geometry or clearance changes; they also face considerable difficulty in sub-millimeter insertion. IL is sample-efficient but performance is constrained by demonstration coverage; under tight tolerances, accumulated errors can readily cause failures, and IL policies typically lack the ability to recover from failures. RL can learn contact strategies and corrective behaviors through interaction, but in real environments it often suffers from low sample efficiency and slow convergence, and may exhibit unsafe exploration. Overall, how to simultaneously achieve learning efficiency, position generalization, failure recovery, and low contact forces under tight tolerances remains an open challenge.

Human operators typically perform grasping and approaching in an open-loop manner using vision, and activate tactile feedback for real-time correction once contact occurs. Inspired by this observation, we propose a two-stage IL+RL framework, as illustrated in Fig.~\ref{fig:teaser}, which decomposes the peg-in-hole task into a vision-guided reach stage and a tactile-augmented insert stage. The reach stage uses IL to learn a grasp-and-approach policy that generalizes over randomized peg initial positions within a $5 \times 5$\,cm workspace, thereby bringing the peg to a local insertion neighborhood and reducing the exploration space for real-robot RL. The insert stage then uses RL to learn contact-rich insertion and corrective behaviors under tight tolerances. To make tactile feedback useful under grasp-dependent offsets, we introduce a residual tactile critic that evaluates actions using insertion-induced tactile changes rather than raw tactile readings. We further propose tactile group sampling, which balances replay updates across post-grasp tactile groups and improves coverage of under-represented contact conditions.

The main contributions of this work are threefold: (i) we formulate a realistic grasp-to-insert precision assembly setting in which the robot starts from a randomly placed peg and performs tight-tolerance insertion after autonomous grasping; (ii) we develop a tactile-aware real-robot RL scheme with tactile baseline compensation and a residual tactile critic for stable value estimation during contact-rich insertion; and (iii) we introduce tactile group sampling to mitigate imbalanced replay distributions caused by different grasp configurations. Systematic experiments across five hole geometries and three clearance settings show that the proposed method consistently improves insertion performance; under the most challenging 0.05\,mm clearance, it achieves a 67\% success rate while maintaining low contact forces and demonstrating position generalization over the randomized $5 \times 5$\,cm workspace.

\section{RELATED WORK}
\subsection{Model-based Insertion Strategies}
Early studies on insertion focused on analytic contact modeling, where force/torque equations were used to classify contact states. Visual pipelines were further incorporated to obtain pose information and estimate alignment errors for motion planning and corrective control \cite{2,3}.
However, such approaches typically require accurate object models and contact assumptions, and are often constrained to specific object types and assembly conditions \cite{4}.
To achieve compliant and safe insertion, many methods employ variable impedance control \cite{5} or combine visual servoing with force/impedance control \cite{6,6-1} to regulate interaction forces while maintaining alignment accuracy.
Other works aim to reduce the reliance on precise perception by proposing manipulation funnel formulations \cite{7,8} or optimization-based contact search strategies \cite{9}.
Overall, model-based strategies are practical and interpretable, but their robustness is often limited by modeling errors, manual tuning costs, and unavoidable pose/contact uncertainties in real-world settings.

\subsection{Learning-based Insertion Strategies}
\begin{figure}[t]
	\centering
	\includegraphics[width=\columnwidth]{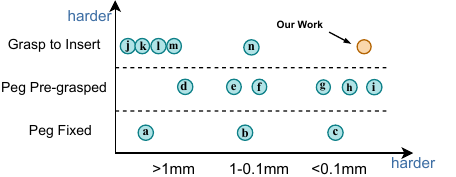}
	\caption{
		Overview of prior peg-in-hole works categorized by task setting (peg fixed, pre-grasped, grasp-to-insert) and tolerance regime ($>1$\,mm, $0.1$--$1$\,mm, $<0.1$\,mm). 
		The literature is mapped as: 
		a:~\cite{18}, b:~\cite{3}, c:~\cite{5}, d:~\cite{38}, e:~\cite{23}, f:~\cite{12}, g:~\cite{34}, h:~\cite{20}, i:~\cite{29}, j:~\cite{30}, k:~\cite{35}, l:~\cite{27}, m:~\cite{21}, n:~\cite{24}. 
		The proposed work targets the challenging grasp-to-insert setting under tight tolerance ($c < 0.1$\,mm, specifically $c = 0.05$\,mm).
	}
	\label{fig:example}
\end{figure}
Some works leverage deep neural networks to estimate 6D poses and contact states \cite{10,11,12,13}, improving robustness but largely retaining a pipeline structure that depends on downstream planning and control.
Imitation learning methods range from lightweight policies such as diffusion-based approaches \cite{14,15} to large-scale models such as VLA-style policies \cite{16,17}.
However, pure imitation learning often struggles to acquire recovery behaviors: small errors can accumulate during contact-rich interactions, easily pushing the policy into out-of-distribution (OOD) states and causing failures.
Consequently, many works focus on reinforcement learning (RL) to address tight-tolerance insertion \cite{18,20,22}, either by training policies in simulation and transferring them to the real world \cite{21,24}, or by performing online training directly on real robots \cite{25,26}.
Given that tight clearances can lead to large interaction forces and safety risks, some studies incorporate tactile feedback into policy pretraining and fine-tune with RL \cite{27}.
To improve sample efficiency and safety, residual RL uses classical controllers or demonstrated behaviors as a base policy and learns only residual corrections \cite{28}.
Force-guided exploration further constrains exploration by limiting interaction forces, promoting force-bounded behaviors, yet often suffers from low exploration efficiency \cite{23}.
While demo-augmented RL improves sample efficiency \cite{29}, neglecting the imbalanced and phase-dependent nature of tactile signals often leads to training instability in contact-rich regimes.
\subsection{Tactile Sensing for Contact-Rich Manipulation}
During contact, relying solely on vision or pose feedback is often insufficient to distinguish subtle misalignments and contact modes. Therefore, many works use force/torque sensing \cite{6,23} or tactile observations \cite{27,30} for contact-state estimation \cite{10,31}.
Others exploit active interactions to model the environment and contact relations \cite{11,32}, and tactile feedback is also frequently used as a core signal for high-frequency reactive control \cite{33}.

\begin{figure*}[t]
	\centering
	\includegraphics[width=0.75\textwidth]{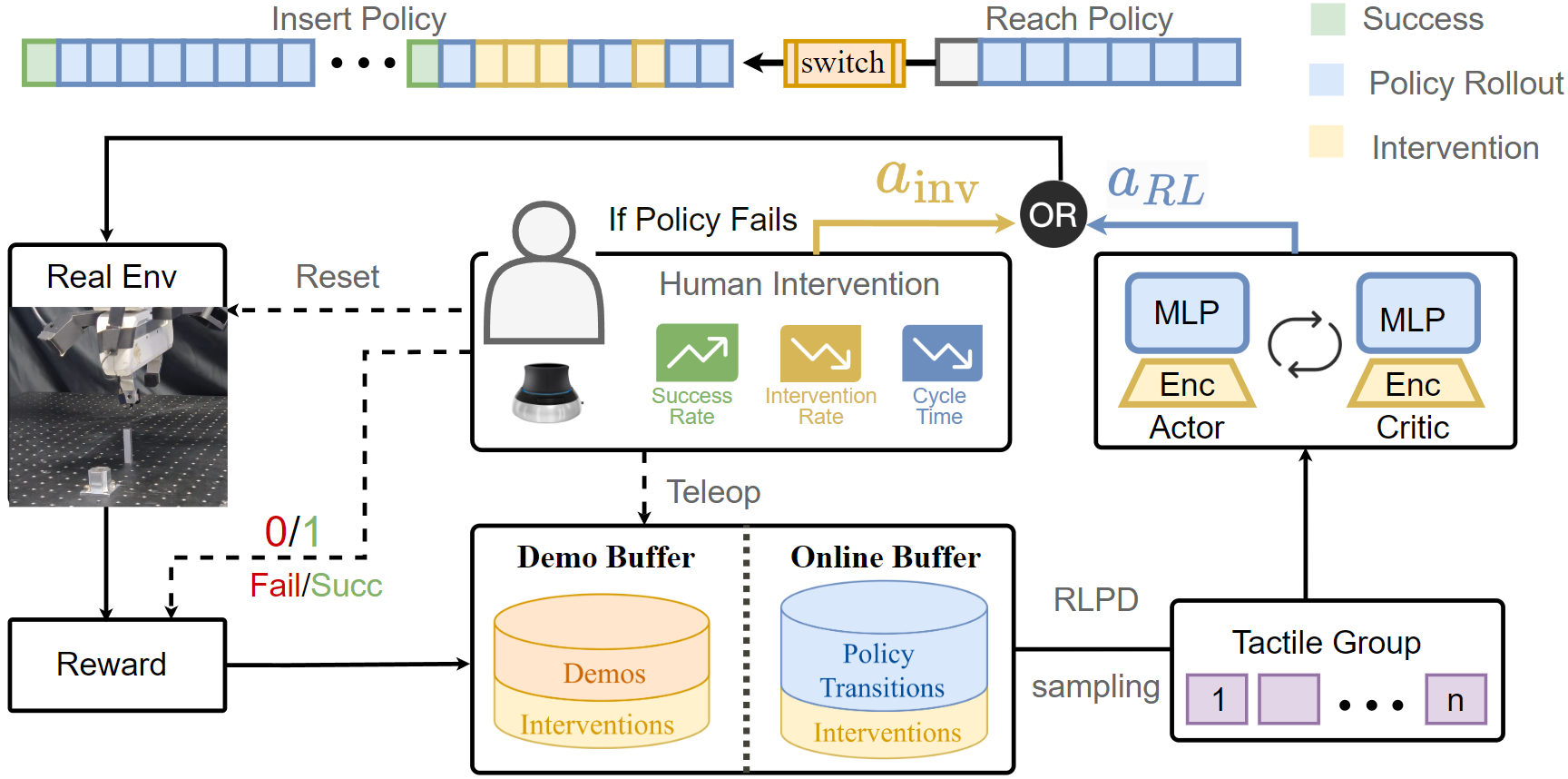}
	\caption{
		Overview of the proposed two-stage framework for peg-in-hole insertion. 
		In the \textit{Reach} stage, a diffusion policy generates short-horizon action sequences 
		to move the peg near the hole entrance. In the \textit{Insert} stage, 
		Soft Actor-Critic (SAC) learns contact-rich insertion behaviors and recovery strategies. 
		Human-in-the-loop intervention enables continuous collection of high-value interaction data, 
		accelerating policy convergence.
	}
	\label{fig:method}
\end{figure*}
With the development of learning-based methods, an increasing number of works incorporate tactile signals as additional observations in imitation learning \cite{34,34-1} or RL policies \cite{35,18}.
A common direction is visuo-tactile fusion: vision provides global geometry and task priors, while touch provides local contact and force details, improving alignment accuracy and robustness in contact-rich phases \cite{36,37}.
Recent representation learning approaches further process raw tactile observations to obtain more stable tactile features, improving policy learning stability and generalization \cite{39,40}.

\section{Problem Formulation}
We study a grasp-to-insert peg-in-hole assembly task on a 7-DoF Franka robot equipped with a tactile-sensing gripper. The goal is to safely insert pegs into corresponding holes with diverse hole geometries under tight clearances.
At the beginning of each episode, the peg is randomly placed within a workspace region. The robot must first autonomously grasp the peg and then execute insertion under tight tolerances.
The system has access to multimodal sensory feedback including vision, tactile sensing, and robot proprioception. In the proposed asymmetric design, visual observations are used by the actor for action generation, while tactile changes are incorporated into the critic during reinforcement learning for value estimation.

We denote the diametral clearance by $c>0$, which characterizes the total difference in diameter between the hole and the peg (smaller $c$ indicates a more challenging setting). Tight-tolerance insertion is difficult for two reasons: (i) precise alignment is required to initiate insertion, and (ii) once contact occurs, even slight pose errors can be amplified into large frictional forces and torques, leading to jamming or excessive interaction forces.
Fig~\ref{fig:example} summarizes prior works along two axes: task setting (peg fixed / pre-grasped / grasp-to-insert) and tolerance regime ($>1$\,mm, $0.1$--$1$\,mm, $<0.1$\,mm). Our work targets the challenging regime of grasp-to-insert with $c<0.1$\,mm (specifically $c=0.05$\,mm).

\section{Method}
To achieve safe insertion across diverse hole geometries under tight tolerances, a two-stage framework is adopted as illustrated in Fig. 3. In the Reach stage, a diffusion policy generates a short-horizon action sequence to bring the grasped peg to the vicinity of the hole entrance. In the Insert stage, Soft Actor-Critic (SAC) is utilized to learn contact-rich insertion behaviors, including recovery from failures. We adopt an asymmetric actor--critic design in this stage: the actor remains vision-dominant for action generation, while tactile feedback is introduced in the critic to improve value estimation during contact-rich interactions. Furthermore, human-in-the-loop intervention is introduced, enabling the continuous collection of high-value interaction data and accelerating convergence.

\subsection{Reach policy}
A dedicated reach policy $\pi_{\mathrm{reach}}$ is trained for each hole geometry to guide the peg to the target vicinity following the grasp. The reach stage employs a diffusion policy trained via imitation learning. It takes as input the end-effector pose and multi-view image observations (comprising a wrist-mounted camera and a global camera) and outputs a short-horizon action sequence executed in a receding-horizon manner:
\begin{equation}
	a_{t:t+H-1}\sim \pi_{\mathrm{reach}}(\cdot \mid o^{\mathrm{reach}}_t).
\end{equation}
By randomizing the peg's initial position within a defined area ($5 \times 5$\,cm in this work), the policy acquires robust positional generalization capabilities. This stage reliably brings the peg into a local insertion neighborhood, thereby substantially reducing the exploration space for the subsequent contact-rich insertion stage.

\subsection{Insert policy}
The insertion stage is modeled as a Partially Observable Markov Decision Process (POMDP) $\mathcal{M}=\{\mathcal{O},\mathcal{A},\rho,\mathcal{P},r,\gamma\}$, where $o_t\in\mathcal{O}$ denotes the multimodal observation, $a_t\in\mathcal{A}$ the action, $\rho(o_0)$ the initial-state distribution, $\mathcal{P}$ the unknown transition dynamics, $r$ the reward function, and $\gamma$ the discount factor. A sparse, manually labeled binary reward is employed:
\begin{equation}
	r_t=
	\begin{cases}
		1, & \text{if insertion succeeds},\\
		0, & \text{otherwise}.
	\end{cases}
\end{equation}
The optimal policy maximizes the expected discounted return: $\max_{\pi} \mathbb{E}[\sum_{t=0}^{T}\gamma^t\,r_t]$.

Soft Actor-Critic (SAC) is adopted as the underlying off-policy reinforcement learning algorithm. SAC learns a parametric Q-function $Q_\phi(o,a)$ and a stochastic policy $\pi_\theta(a\mid o)$ with entropy regularization. The critic loss is given by:
\begin{equation}
	\mathcal{L}_Q(\phi)=
	\mathbb{E}_{(o,a,r,o',d)}\Big[
	\big(Q_{\phi}(o,a)-y\big)^2
	\Big],
\end{equation}
where the target value is
\begin{equation}
	y=r+\gamma(1-d)\,
	\mathbb{E}_{a'\sim\pi_{\theta}(\cdot\mid o')}
	\Big[
	Q_{\bar{\phi}}(o',a')-\alpha\log \pi_{\theta}(a'\mid o')
	\Big],
\end{equation}
$\bar{\phi}$ denotes the target critic parameters, $\alpha$ is the temperature coefficient, and $d \in \{0, 1\}$ represents the terminal flag indicating whether the episode has ended (e.g., successful insertion). The actor loss is
\begin{equation}
	\mathcal{L}_{\pi}(\theta)=
	\mathbb{E}_{o}\Big[
	\mathbb{E}_{a\sim\pi_{\theta}(\cdot\mid o)}
	\big[\alpha \log \pi_{\theta}(a\mid o)-Q_{\phi}(o,a)\big]
	\Big].
\end{equation}

To improve sample efficiency on the real robot, an RLPD-style \cite{26} training pipeline is adopted. Two replay buffers are maintained: a demonstration/prior buffer $\mathcal{D}_{\mathrm{demo}}$ and an online buffer $\mathcal{D}_{\mathrm{on}}$. Each update samples approximately equally from the two buffers to form a mini-batch, and the SAC losses are used to update $Q_\phi$ and $\pi_\theta$. In the human-in-the-loop setting, if the policy enters an unrecoverable or high-risk state, an operator takes over to provide corrective actions $a^{\mathrm{inv}}$. These transitions are stored in $\mathcal{D}_{\mathrm{demo}}$ to stabilize training and accelerate convergence.

During the insertion stage, global observations are replaced with an object-centric cropped view focused on the peg and hole region. Since the exploration space is restricted during insertion, visual variance is primarily concentrated in this area; thus, cropping effectively removes global redundant information. Additionally, centering the hole within the cropped frame helps the visual encoder better represent the insertion process, maintaining policy stability even when the global hole position varies.

\begin{algorithm}[t]
	\caption{Two-stage training with IL and RL}
	\label{alg:two_stage}
	\begin{algorithmic}[1]
		\STATE \textbf{Input:} Reach demos $\mathcal{D}^{\mathrm{reach}}_{\mathrm{demo}}$, Insertion demos $\mathcal{D}^{\mathrm{ins}}_{\mathrm{demo}}$
		\STATE Initialize $\pi_{\mathrm{reach}}$, $\pi_{\mathrm{insert}}$, buffers $\mathcal{D}_{\mathrm{demo}} \leftarrow \mathcal{D}^{\mathrm{ins}}_{\mathrm{demo}}$, $\mathcal{D}_{\mathrm{on}} \leftarrow \emptyset$
		\STATE Extract demo endpoints $\mathcal{P}_{\mathrm{demo}}$; set $\mathcal{P}_{\mathrm{fail}} \leftarrow \emptyset$; build $\mathcal{X}_{\mathrm{init}} \leftarrow \mathrm{AABB}(\mathcal{P}_{\mathrm{demo}} \cup \mathcal{P}_{\mathrm{fail}})$
		\STATE \textbf{Critic Warmup:} Pretrain $Q_\phi$ on $\mathcal{D}_{\mathrm{demo}}$ via Monte Carlo (MC) target (Eq. \ref{eq:mc_loss})
		\STATE Train $\pi_{\mathrm{reach}}$ on $\mathcal{D}^{\mathrm{reach}}_{\mathrm{demo}}$ via Diffusion Policy
		\WHILE{not converged}
		\STATE Execute $\pi_{\mathrm{reach}}$ to grasp/approach; record tactile baseline $\bar{\mathbf{f}}$; update $\mathcal{X}_{\mathrm{init}}$
		\STATE Switch to $\pi_{\mathrm{insert}}$ when $\mathbf{p} \in \mathcal{X}_{\mathrm{init}}$
		\FOR{each insertion attempt}
		\STATE Reset $\mathbf{p}_0 \sim \mathrm{Unif}(\mathcal{P}_{\mathrm{fail}})$ if $\mathcal{P}_{\mathrm{fail}} \neq \emptyset$
		\FOR{$t=0$ to $T-1$}
		\STATE Execute $a_t^{\mathrm{RL}} \sim \pi_{\mathrm{insert}}(\cdot \mid o_t)$ (or $a_t^{\mathrm{inv}}$ if intervened); store to $\mathcal{D}_{\mathrm{on}}$ and $\mathcal{D}_{\mathrm{demo}}$ (if intervened)
		\STATE Every $U$ steps, Train $\pi_{\mathrm{insert}}$ with Tactile Group Sampling from $\mathcal{D}_{\mathrm{demo}}$ and $\mathcal{D}_{\mathrm{on}}$
		\ENDFOR
		\STATE On failure, append endpoint to $\mathcal{P}_{\mathrm{fail}}$ and update $\mathcal{X}_{\mathrm{init}}$
		\ENDFOR
		\ENDWHILE
	\end{algorithmic}
\end{algorithm}

\paragraph{Location-based two-stage switching}
During both training and inference, a location-based switching mechanism is adopted. Let $\mathbf{p}\in\mathbb{R}^3$ denote the end-effector position. The stage indicator is defined as
\begin{equation}
	I_{\mathrm{ins}}(\mathbf{p})=
	\begin{cases}
		1, & \mathbf{p}\in \mathcal{X}_{\mathrm{init}},\\
		0, & \text{otherwise},
	\end{cases}
\end{equation}
and the execution policy is written as a one-way stage-transition rule:
\begin{equation}
	\pi_{\mathrm{exec}}(a_t\mid o_t)=
	\begin{cases}
		\pi_{\mathrm{reach}}(a_t\mid o_t), & \mathbf{p}_t\notin \mathcal{X}_{\mathrm{init}},\\
		\pi_{\mathrm{insert}}(a_t\mid o_t), & \mathbf{p}_t\in \mathcal{X}_{\mathrm{init}}.
	\end{cases}
\end{equation}
Here, $\mathcal{X}_{\mathrm{init}}$ is defined as the axis-aligned bounding box (AABB) of the union of demo end points $\mathcal{P}_{\mathrm{demo}}$ and valid reach-failure points $\mathcal{P}_{\mathrm{fail}}$: $\mathcal{X}_{\mathrm{init}} = \mathrm{AABB}(\mathcal{P}_{\mathrm{demo}} \cup \mathcal{P}_{\mathrm{fail}})$. To prevent abnormal task terminations from excessively expanding the bounding box, statistical outlier rejection is applied to $\mathcal{P}_{\mathrm{fail}}$ before dynamically updating $\mathcal{X}_{\mathrm{init}}$, thereby accelerating the learning process safely.

\begin{figure}[t]
	\centering
	\includegraphics[width=\columnwidth]{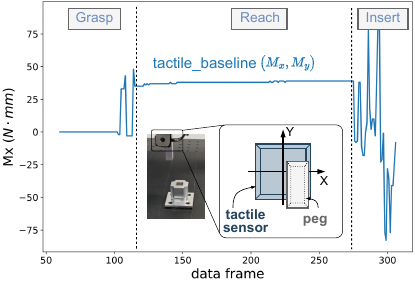}
	\caption{Illustration of the tactile baseline estimation and the resulting tactile change signals across different grasp poses.}
	\label{fig:tactile}
\end{figure}

\paragraph{Tactile Critic}
To compensate for grasp-dependent tactile offsets, a tactile baseline $\bar{\mathbf{f}}$ is estimated after each grasp, as shown in Fig~\ref{fig:tactile}. Specifically, tactile readings are averaged over a short stable window following gripper closure:
\begin{equation}
	\bar{\mathbf{f}}=\frac{1}{K}\sum_{k=0}^{K-1}\mathbf{f}_{t_b+k},
\end{equation}
where $\bar{\mathbf{f}} = [F_x, F_y, F_z, M_x, M_y, M_z]^\top$ represents the mean 6-DoF force and torque vector, and $t_b$ denotes the start index of the stable post-grasp interval. The tactile baseline is critical for characterizing the grasp state; variations in the grasp position lead to significant discrepancies in the raw sensor signals, particularly for torques $M_x$ and $M_y$. Direct input of raw tactile readings into the critic would introduce excessive variance across different grasp configurations, hindering the policy's ability to identify effective tactile patterns and impeding convergence. Consequently, the tactile change is defined as $\Delta\mathbf{f}_t=\mathbf{f}_t-\bar{\mathbf{f}}$, which isolates insertion-induced contact variations.

An asymmetric actor--critic architecture is adopted, where the actor and critic share a common ResNet visual encoder. The actor relies solely on visual observations for action generation, while the critic additionally incorporates tactile changes $\Delta f_t$ for value estimation. This design is motivated by the fact that tactile signals are strongly affected by grasp-dependent offsets and can vary significantly across different in-hand peg poses. Directly feeding such signals into the actor may increase input variance and destabilize policy learning. Instead, tactile feedback is used in the critic to provide contact-aware value estimation, enabling the actor to learn stable visually grounded insertion behaviors. To provide a structured initial value surface for RL, a Critic Warmup stage is implemented using the demonstration buffer $\mathcal{D}_{\mathrm{demo}}$. In sparse-reward insertion tasks, a Monte Carlo (MC) loss is utilized to directly regress the critic toward the expert's cumulative returns $G$, bypassing the bootstrap error inherent in early-stage temporal difference learning:
\begin{equation}
	\label{eq:mc_loss}
	\mathcal{L}_{\mathrm{MC}}(\phi) = \mathbb{E}_{(s,a,G) \sim \mathcal{D}_{\mathrm{demo}}} \left[ \left( Q_\phi(s, a) - G \right)^2 \right],
\end{equation}
where $G$ represents the pre-computed Monte Carlo returns associated with demonstration trajectories. This supervised initialization ensures that the critic captures the reward structure before interaction begins, significantly accelerating policy convergence.

\paragraph{Tactile Group Sampling}
Off-policy training may lead to imbalanced replay distributions dominated by specific grasp configurations, reducing robustness across diverse in-hand poses. To improve the coverage of contact-relevant modes, transitions are grouped according to the baseline torque components $(\bar{M}_x,\bar{M}_y)$ recorded after grasping.

The components $\bar{M}_x$ and $\bar{M}_y$ are discretized into $N$ and $M$ intervals, respectively, where the bin boundaries are determined based on the empirical range of the demonstration data. This results in $N \times M$ tactile groups. The group index is defined as:
\begin{equation}
	g(\bar{\mathbf{f}})=
	\big(b_x(\bar{M}_x),\, b_y(\bar{M}_y)\big),
\end{equation}
where $b_x(\cdot)$ and $b_y(\cdot)$ denote the discretization functions along the $M_x$ and $M_y$ axes. During training, groups are sampled approximately uniformly, followed by uniform sampling within the selected group. This grouped replay strategy increases the update frequency of under-represented grasp conditions, thereby improving stability and generalization.

\begin{table*}[t]
	\centering
	\caption{Success rates of different methods across varying hole geometries and assembly clearances.}
	\label{tab:success_rate_comparison}
	\small
	\setlength{\tabcolsep}{12pt} 
	
	\begin{adjustbox}{max width=\textwidth}
		\begin{tabular}{ll ccccc}
			\toprule
			Method & Clearance & Square & Round & Hexagonal & L-shape & Triangular \\
			\midrule
			\multirow{3}{*}{IL} 
			& 1\,mm  & 10/15 & 13/15 & 10/15 & 11/15 & 13/15 \\
			& 0.25\,mm & 4/15  & 2/15  & 4/15  & 2/15  & 6/15  \\
			& 0.05\,mm & 0/15  & 0/15  & 0/15  & 0/15  & 0/15  \\
			\cmidrule(lr){1-7}
			
			\multirow{3}{*}{IL + RL} 
			& 1\,mm  & 15/15 & 14/15 & 14/15 & 15/15 & 15/15 \\
			& 0.25\,mm & 12/15 & 8/15  & 12/15 & 13/15 & 14/15 \\
			& 0.05\,mm & 8/15  & 6/15  & 8/15  & 5/15  & 12/15 \\
			\cmidrule(lr){1-7}
			
			\multirow{3}{*}{\makecell[l]{IL + RL with tactile critic\\and tactile group}} 
			& 1\,mm  & 15/15 & 14/15 & 14/15 & 15/15 & 15/15 \\
			& 0.25\,mm & 14/15 & 14/15 & 14/15 & 14/15 & 15/15 \\
			& 0.05\,mm & \textbf{11/15} & \textbf{7/15}  & \textbf{10/15} & \textbf{8/15}  & \textbf{14/15} \\
			\bottomrule
		\end{tabular}
	\end{adjustbox}
\end{table*}
\begin{figure}[t]
	\centering
	\includegraphics[width=0.7\columnwidth]{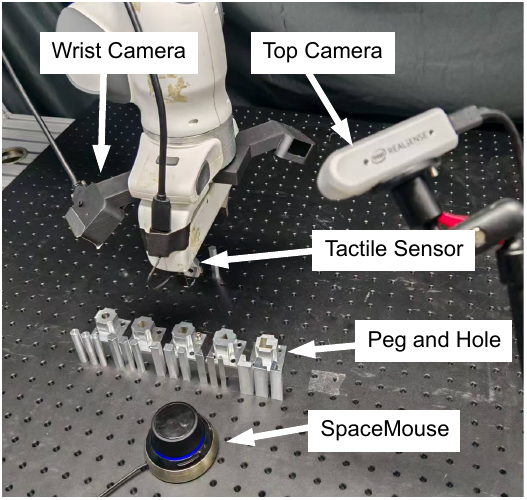}
	\caption{Experimental platform overview.}
	\label{fig:experiment}
\end{figure}
\section{Experiments}

\subsection{Experimental Setup}
As shown in Fig.~\ref{fig:experiment}, the system features a 7-DOF Franka Emika Panda robotic arm equipped with a Paxini PX-6AX GEN3 tactile sensor capable of providing high-precision six-axis force/torque measurements. For perception, a wrist-mounted Intel RealSense D405 camera enables fine-grained local observation, while a RealSense D435i camera supplies global scene awareness. Human demonstrations and corrective interventions are facilitated through a 3Dconnexion SpaceMouse interface. For computation, a workstation equipped with an NVIDIA RTX 5090 GPU and an Intel Core i9-14900 CPU is utilized for model training and real-time inference. The assembly task involves CNC-machined metal peg-in-hole workpieces with five geometries (square, round, hexagonal, L-shape, and triangular) and three clearances (0.05, 0.25, and 1\,mm). During each trial, the hole is securely fixed, and the peg's initial position is randomized within a $5 \times 5\,\mathrm{cm}$ area. The robot is tasked to autonomously grasp the peg and execute the insertion.

\subsection{Implementation Details}
\textbf{Reach Stage Policy:} For the reach stage, a diffusion policy is trained using 100 human demonstrations for each geometry. The policy is trained for 200 epochs with a batch size of 64, employing a ResNet-18 encoder to process $128 \times 128$ image observations. The action horizon is set to $H=16$ with an observation horizon of $T=2$, enabling smooth reaching trajectories to the neighborhood of the hole.

\textbf{Real-Robot Fine-tuning:} During the insertion stage, the policy is fine-tuned using Soft Actor-Critic (SAC) with a pixels-based observation space. An RLPD-style training pipeline is employed, sampling approximately equally from the online buffer and the demonstration buffer. To address high-precision contact challenges, an extra critic update ratio (CTA ratio) is applied, and the tactile group sampling is performed across $2 \times 4$ intervals of the post-grasp torques $\bar{M}_x$ and $\bar{M}_y$. The control frequency is maintained at 10\,Hz, and zero-intervention success is typically achieved within 30 minutes of human-in-the-loop training, totaling approximately 1,000 to 2,000 environment transitions.

\subsection{Success Rate and Comparative Analysis}
The proposed framework is evaluated against two baselines:
\begin{itemize}
	\item \textbf{Pure IL:} Employs imitation learning for both reach and insert stages.
	\item \textbf{IL + RL:} Uses IL for reaching and SAC-based RL for insertion, but without tactile observations.
	\item \textbf{Proposed (Full):} The complete IL + RL framework with the tactile-augmented critic and tactile group sampling.
\end{itemize}

\begin{figure}[t]
	\centering
	\includegraphics[width=\columnwidth]{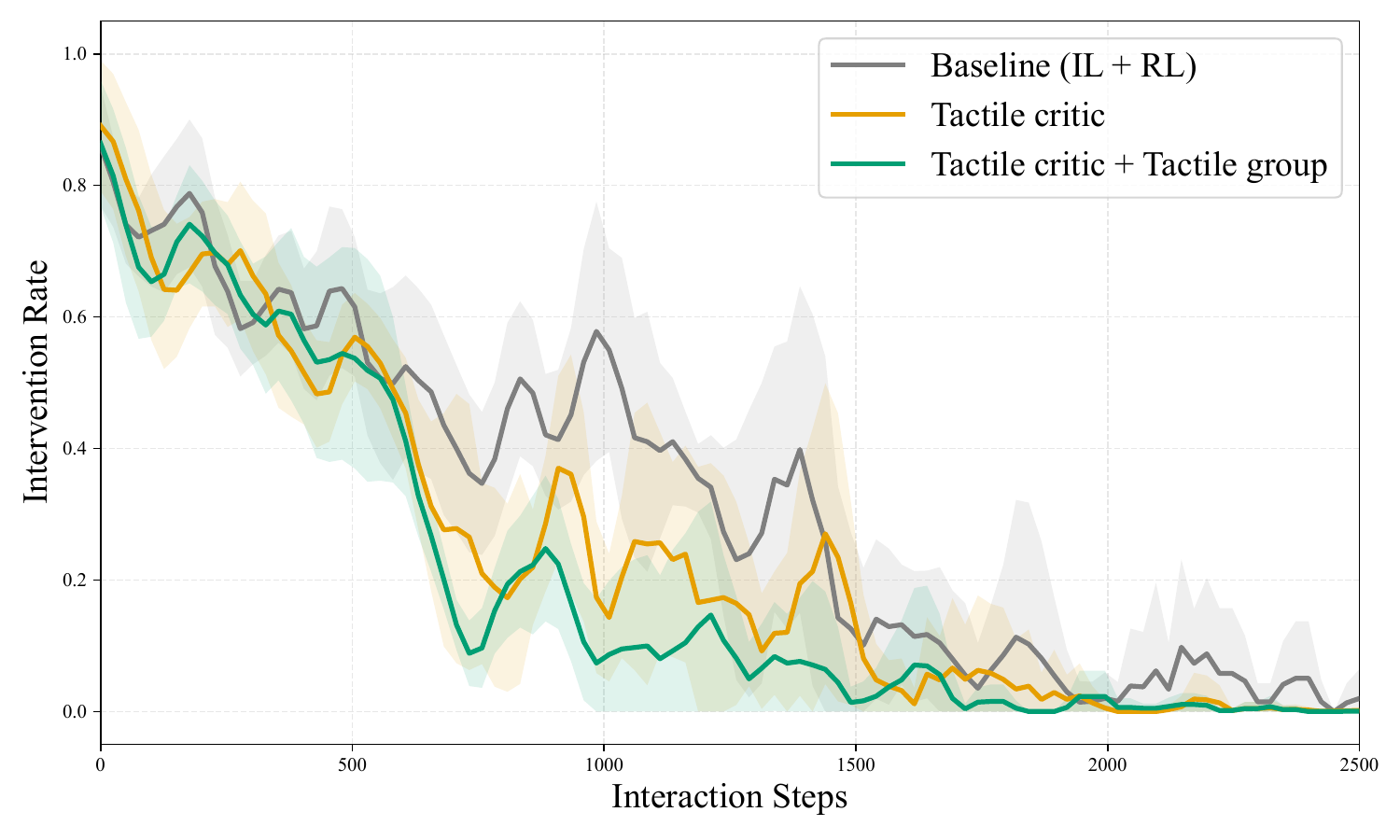}
	\caption{Comparison of human intervention rates during training for the 0.05\,mm clearance task. The solid lines represent the mean values, while the shaded regions denote the standard deviation across multiple independent runs.}
	\label{fig:training_analysis}
\end{figure}
As shown in Table~\ref{tab:success_rate_comparison}, all methods perform well at the $1$\,mm clearance level. Pure IL achieves an average success rate of 76\%, while RL-based methods reach nearly 100\%. In this regime, large interaction forces are rare, allowing IL to succeed with high-quality demonstrations. However, in the sub-millimeter regime ($0.25$\,mm and $0.05$\,mm), Pure IL fails significantly (success rates drop to 24\% and 0\%, respectively) due to pose error accumulation and the lack of recovery behaviors. Specifically, high interaction forces cause the peg to shift within the gripper, leading to unrecoverable drifts.

In contrast, RL-based methods allow the policy to learn from failures and master exploratory behaviors. After only 4,000 interaction transitions, the vanilla IL+RL improves success rates to 79\% ($0.25$\,mm) and 52\% ($0.05$\,mm). The proposed full method achieves the best performance, with success rates further rising to 95\% and 67\%, respectively, demonstrating the efficacy of the tactile critic and tactile group sampling in handling tight tolerances.
\begin{figure}[t]
	\centering
	\includegraphics[width=0.98\columnwidth]{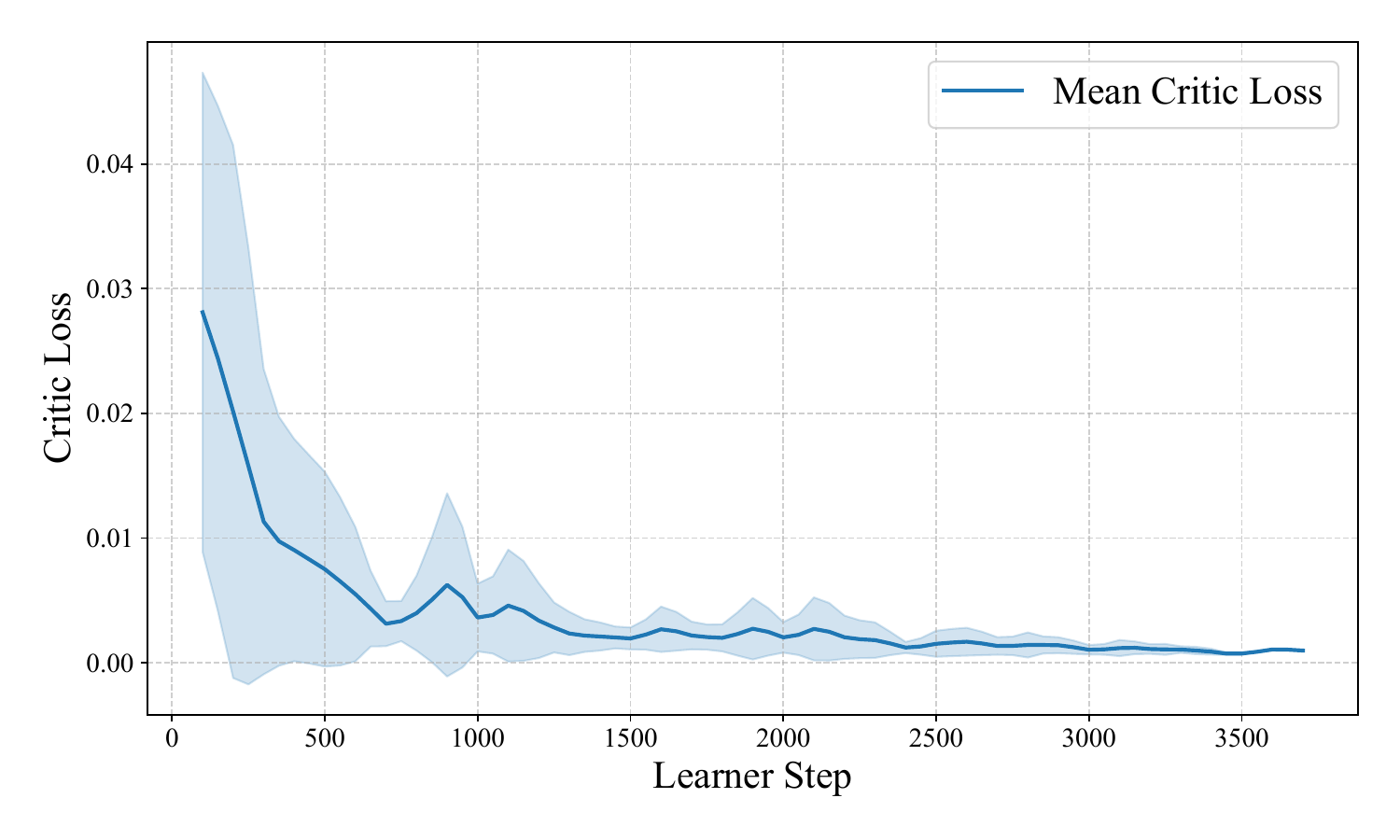} 
	\caption{Aggregated critic loss during online reinforcement learning. The solid line represents the smoothed mean loss, while the shaded region denotes the standard deviation.}
	\label{fig:critic_loss_convergence}
\end{figure}
\begin{figure}[t]
	\centering
	\includegraphics[width=0.95\columnwidth]{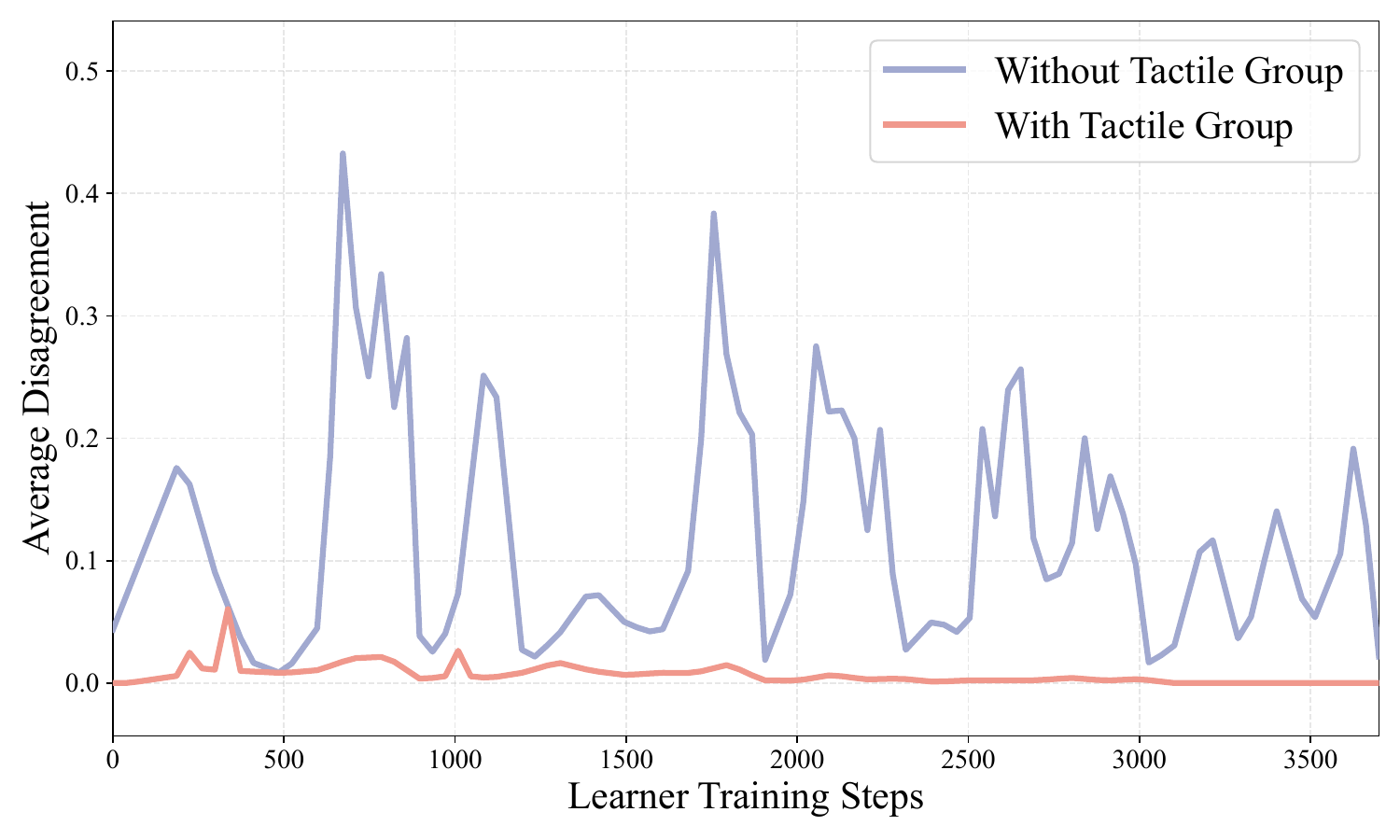}
	\caption{
		Comparison of Q-value disagreement during training.
		The proposed Tactile Group Sampling exhibits consistently lower
		Q-disagreement than the baseline, indicating more stable value estimation.
	}
	\label{fig:q_disagreement}
\end{figure}

\begin{figure*}[t]
	\centering
	\includegraphics[width=0.9\textwidth]{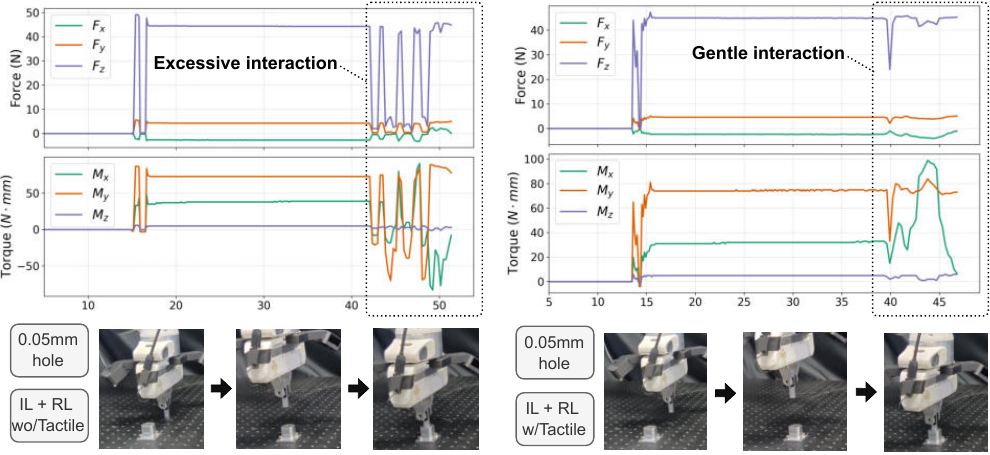}
	\caption{
		Interaction force and torque comparison results between vanilla IL+RL 
		and the proposed method during real-world 0.05\,mm insertion. 
		Vanilla policies exhibit high-frequency, high-amplitude force spikes, with peak normal force fluctuations exceeding 50\,N and torque variations surpassing 160\,N$\cdot$mm. 
		In contrast, the proposed method regulates interaction forces within a safe range, limiting the maximum normal force fluctuation to 20\,N and torque variation to 90\,N$\cdot$mm.
	}
	\label{fig:force_analysis}
\end{figure*}
\subsection{Learning Convergence and Efficiency Analysis}

The effectiveness of the training process is evaluated through both the convergence of the tactile critic and the reduction in human intervention. During the pre-training phase, the critic undergoes a supervised warmup to ground the action-value function in the expert return surface. The critic loss diminishes from an initial 0.503 to approximately 0.002 within 160 gradient steps, while the predicted $Q_{\text{mean}}$ aligns closely with the expert Monte Carlo returns ($G \approx 0.94$). This structured initialization provides a reliable value prior, mitigating bootstrap errors typically associated with early-stage sparse-reward reinforcement learning and providing a stable foundation for online interaction.

In the subsequent online RL phase, the training trajectory is analyzed as illustrated in Fig.~\ref{fig:training_analysis}. While vanilla IL+RL eventually achieves zero-intervention insertion, the intervention rate fluctuates significantly due to poor robustness against varying in-hand poses. The tactile-augmented critic allows the intervention rate to converge more stably, enabling zero-shot adaptation to new grasp configurations without additional training. 

As further evidenced by the aggregated training curves in Fig.~\ref{fig:critic_loss_convergence}, the critic loss is consistently maintained within a stable range of $10^{-3}$ to $10^{-2}$ throughout the online fine-tuning process. Such low-magnitude loss and stable gradients are essential for the actor to master precise corrective behaviors under stringent 0.05\,mm clearance constraints.

Furthermore, \textit{Tactile Group Sampling} significantly accelerates learning efficiency, achieving the first successful zero-intervention insertion at approximately 1,000 transitions. The training procedure inherently leads to a sample distribution that is initially concentrated on a limited set of grasp configurations. Grouped sampling ensures a uniform distribution of contact modes within each replay mini-batch, thereby stabilizing the Q-value estimation. This is further evidenced in Fig.~\ref{fig:q_disagreement}, where the proposed strategy exhibits lower Q-disagreement compared to the baseline, demonstrating the efficacy of balancing the replay distribution for consistent value estimation across diverse contact regimes.

\subsection{Safety and Interaction Force Analysis}
Finally, the safety of the proposed method is validated during real-world $0.05$\,mm insertions. We further evaluate the impact of incorporating tactile feedback on interaction dynamics. Fig.~\ref{fig:force_analysis} illustrates the interaction force and torque profiles. While vanilla IL+RL policies achieve successful insertions, they trigger high-frequency, high-amplitude force spikes during contact, with peak normal force fluctuations exceeding 50\,N and torque variations surpassing 160\,N$\cdot$mm. In contrast, the proposed method regulates interaction forces within a safe range, limiting the maximum normal force fluctuation to 20\,N and torque variation to 90\,N$\cdot$mm, corresponding to reductions of 60\% and 44\%, respectively. By proactively detecting contact states via the tactile critic, the policy learns to suppress excessive forces, ensuring both a high success rate and operational safety in precision assembly tasks.

\section{CONCLUSIONS}

This paper presents a tactile-augmented two-stage imitation and reinforcement learning framework for high-precision robotic assembly under tight tolerances. The introduction of tactile group sampling significantly improves training stability by balancing the experience replay distribution, while the tactile critic ensures more accurate action-value estimation during contact. Real-world experiments across diverse hole geometries and clearances demonstrate that the proposed method achieves superior success rates and maintains low interaction forces. Specifically, for the most challenging 0.05\,mm clearance, the system achieves a 67\% success rate within only 30 minutes of human intervention, while reducing peak normal force and torque by 60\% and 44\%, respectively. Future work will explore the extension of this tactile-augmented framework to multi-part assembly sequences and insertion tasks with even more stringent tolerances.


\balance

\end{document}